\documentclass[letterpaper, 10pt, conference]{IEEEtran}
\IEEEoverridecommandlockouts

\usepackage[letterpaper, left=0.7in, right=0.7in, top=0.8in, bottom=0.9in]{geometry}

\usepackage{pifont} 

\usepackage{graphicx}
\usepackage{xspace}
\usepackage{colortbl}
\usepackage{hyperref}
\usepackage{float}
\usepackage{multirow}
\usepackage{tabularx}
\usepackage[fleqn]{amsmath}
\usepackage{siunitx}
\usepackage{amssymb}

\usepackage[table, xcdraw]{xcolor}

\usepackage{makecell}

\definecolor{lightgray}{gray}{0.95}

\usepackage{subcaption}

\newcommand{\ie}{\emph{i.e.,}\xspace}
\newcommand{\eg}{\emph{e.g.,}\xspace}

\newcommand{\final}[1]{\textcolor{black}{#1}}

\begin{document}


%

\title{\LARGE \bf Tuning ROS 2 for Energy-Efficient Navigation: Empirical Insights from Costmap 2D Configurations}


\author{\IEEEauthorblockN{Michel Albonico\IEEEauthorrefmark{1}\IEEEauthorrefmark{2},
Andreas Wortmann\IEEEauthorrefmark{3}, and
Ivano Malavolta\IEEEauthorrefmark{4}} 
\IEEEauthorblockA{\IEEEauthorrefmark{1}Federal University of Technology - Paraná (UTFPR), DAINF-FB, Francisco Beltrão, Brazil}
\IEEEauthorblockA{\IEEEauthorrefmark{2}University of Southern Denmark (SDU), MMMI, Centre for Software Technology (CST), Vejle, Denmark}
\IEEEauthorblockA{\IEEEauthorrefmark{3}Institute for Control Eng. of Machine Tools and Manufacturing Units (ISW), University of Stuttgart, Stuttgart, Germany
}
\IEEEauthorblockA{\IEEEauthorrefmark{4}Vrije Universiteit (VU) Amsterdam, Software and Sustainability (S2) Group, Amsterdam, The Netherlands
}
\IEEEauthorblockA{michelalbonico@utfpr.edu.br, mical@mmmi.sdu.dk, andreas.wortmann@isw.uni-stuttgart.de, i.malavolta@vu.nl}
}

\maketitle

\begin{abstract}
Robots are increasingly used in diverse application areas, where autonomous navigation plays a central role.
As these systems become more widespread, improving their energy efficiency is critical to extending operational time and reducing environmental impact.
The Robot Operating System (ROS) is a widely adopted middleware for robotics, offering a rich set of configurable packages.
However, this flexibility can result in suboptimal software configurations in dynamic environments, negatively affecting both performance and energy consumption.
This paper investigates the impact of ROS 2 package re-configurations on the energy efficiency of mobile robot navigation.
We conduct a controlled experiment in two warehouse-like scenarios (small and large) with varying obstacle layouts and Costmap 2D configurations (essential to the Nav2 stack).
Through repeated trials, we measure energy usage, power profile, CPU load, memory consumption, and navigation performance.
Results show that configurations must be carefully chosen for the specific robotic environment, and we were able to identify critical settings that lead to good and poor performance and energy consumption.
\end{abstract}


\section{Introduction}

Robots are increasingly deployed across diverse domains, \eg logistics, healthcare, agriculture, where energy efficiency is becoming an essential concern.  
As the demand for autonomous navigation grows, there is a need to reduce energy consumption, both for extending operational time (and consequently, increasing safety) and for minimizing environmental impact~\cite{hertwich2018growing}.  
Among the various factors influencing robots' energy efficiency, software components play a central role, from low-level control to high-level decision-making~\cite{chinnappan2021architectural,9463077,santos2020}.  
In particular, robotics software can itself be a major source of energy consumption~\cite{swanborn2020energy}, making it a key factor in the design of sustainable and energy-aware robotic systems.

ROS has emerged as the de-facto standard for robotics software development~\cite{malavolta2021mining}, offering a wide range of highly configurable packages.  
This extensive configuration space, combined with dynamic robotic environments, can lead to suboptimal software configurations~\cite{PEREIRA2021111044}, which consequently affect both performance and energy efficiency~\cite{albonico2023, hamer2025resource}.  
In this context, setting ROS configurations to specific workloads becomes a promising area for improvement~\cite{xie2019}.

The \textbf{main goal} of this study is to empirically assess the impact of ROS package (re-)configurations on the energy efficiency of mobile robots' navigation stack.  
We build our experiment on the Nav2\footnote{\url{https://docs.nav2.org/}}, the current official navigation package for ROS 2, and propose alternative configurations for the \emph{Costmap 2D} package\footnote{\url{https://docs.nav2.org/configuration/packages/configuring-costmaps.html}}, essential for navigation decision-making.
This study hypotheses is that optimizing the \emph{Costmap 2D} configuration in dynamic environments affects the robot's energy consumption during navigation tasks.

In this work, we conduct a controlled experiment in which a mobile robot autonomously navigates a warehouse floor populated with dynamic obstacles \final{(i.e., objects deterministically added during the mission)}.  
The experiment has four independent variables: the \emph{navigation goal}, the \emph{Costmap 2D configuration}, the \emph{obstacles} arrangement, and the \emph{warehouse layout}.  
The \emph{robot model}, \emph{navigation goal}, and \emph{other ROS package settings} are kept constant throughout the experiment.  
For each run, a fixed navigation goal is sent to the robot, a specific \emph{Costmap 2D} configuration is applied, and the obstacle scenario is set deterministically.  
Each configuration was evaluated over \num{20}x repetitions, balancing statistical reliability and experimental feasibility, which keeps the total execution time within practical limits. 
We collect performance and resource-usage metrics, including energy consumption.

Our \textbf{experimental results} demonstrate that tuning \emph{Costmap 2D} parameters can reduce navigation time and power demands, leading to superior overall energy efficiency.
Across both small and large warehouse layouts, certain parameterizations, notably those using a moderate resolution, low update frequencies, and reduced reliance on inflated or unknown-space layers, consistently achieved lower energy usage. 
In contrast, rough resolutions and excessive inflation values led to navigation failures, longer paths, or corrective behaviors, all of which increased energy consumption.

\section{ROS Configuration}
\label{subsec:ros-conf}
 
ROS packages tend to be highly configurable, given the dynamic environment where the robots are deployed~\cite{canelas2024understanding}.  
That is, they expose a wide range of parameters, allowing users to customize system behavior without modifying source code.
As a result, users can fine-tune performance for their specific hardware, environment, and application requirements.
This flexibility, while powerful, introduces a significant challenge: the configuration space quickly becomes large and complex, which can lead to suboptimal configurations, compromising performance and resource usage.
%
%
%
%
Formally, the \emph{ROS configuration space} ($C$) of a system $S$ can be defined as the Cartesian product of all valid parameter values across all ROS packages within $S$.
The \emph{ROS system} is defined as \( S = \{ \pi_1, \pi_2, \ldots, \pi_m \} \), 
where each \( \pi_j \) is a package. Each package contributes with a subset of parameters, where \( P = \{ p_1, p_2, \ldots, p_n \} \) is the complete set of parameters across all packages in \( S \). 
For each parameter \( p_i \), let \( D_i \) denote its \emph{domain of admissible values}, which may be finite 
(e.g., \( D_i = \{0,1\} \) for booleans) or infinite (e.g., \( D_i = \mathbb{R} \) for a continuous parameter value, 
or \( D_i = [a,b] \) for an interval). The \emph{ROS configuration space} is defined as 
\( C = \prod_{i=1}^n D_i \). A \emph{configuration} is then an element of this space, 
\( c = (v_1, v_2, \ldots, v_n) \) with \( v_i \in D_i \). Thus, each configuration \( c \in C \) 
represents a complete assignment of admissible values to all parameters in the system \( S \).


\section{Related Work}
\label{sec:related_work}

Regarding the energy efficiency for Robot Operating System (ROS), Malavolta et al.~\cite{malavolta2021mining} mined the ROS ecosystem to identify four green architectural tactics, achieving up to 7.9\% energy savings on real hardware systems. 
Albonico et al.~\cite{albonico12energy} presented a large-scale study on how the choice of programming language influences energy efficiency in ROS node implementations. Dordevic et al.~\cite{dhordjevic2023computation} explored the energy implications of offloading ROS components to edge devices, providing empirical evidence on the trade-offs involved in distributed ROS deployments. 
Hammer et al.~\cite{hamer2025resource} evaluate the impact of different Simultaneous Localization and Mapping (SLAM) packages on the energy consumption of the ROS system.
These studies form a foundation for our work, particularly regarding energy metrics and measurement methodologies. 
They diverge from our specific research focus and objectives since none of them focus on alternative configurations of Nav2 packages towards energy efficiency.

Considering robotic software (re-)configuration,
Brugali~\cite{brugali2020} analyzes architectural impediments to
runtime reconfiguration of ROS and proposes a reconfigurable navigation system, apart from the Nav2 stack.
Peldszus et al.~\cite{peldszus2025software} investigate how robotic software can be customized, bringing empirical evidence on robotic software reconfiguration granularity and tools to support reconfiguration. 
Canelas et al.~\cite{canelas2024} systematically categorized \num{50} types of ROS misconfigurations from developer forum data, revealing that more than half remain unaddressed.
Those works reinforce the importance of a broader empirical investigation regarding ROS configurations, and differ from our work since none of them measure the impact of robotic/ROS software configurations.




    

    
    

\section{Robotic Mission}
\label{sec:robotic-mission}

This paper is based on a robotic mission in which a mobile robot navigates a warehouse floor, always going from its charging station to a fixed navigation goal. 
The mission represents a realistic situation in which the robot autonomously moves toward a shelf to pick an object.
We model two warehouse floor layouts, which we simply call \emph{maps} in the remainder of this document: a customized \emph{small} map measuring $3 \times 4$\,m and a publicly-available \emph{large} one\footnote{\url{https://github.com/belal-ibrahim/dynamic_logistics_warehouse}} ($\approx 10$ times larger).
Both maps contain different furniture, including storage racks, tables, and stacks of boxes that create navigation spaces of varying proportions (\ie, narrow and wide passages). 
The robot begins at a hypothetical charging station, with its navigation goal set at an opposite side of the warehouse, aiming at navigation paths of $\approx6$ (\emph{small} map) and $\approx18$ (\emph{large} map) meters, which triples the distance.
The navigation in the \emph{small} map is constrained in \num{180} seconds, while in the \emph{large} map it is \num{540} seconds (also the triple).
The environment may have \final{obstacles added dynamically}, deterministically and strategically placed so as not to completely block the robot’s route, and in a time window that forces the robot to identify them after the navigation starts.

\section{Study Design}
\label{sec:experiment_definition}



By following the Goal Question Metric (GQM) framework~\cite{van2002goal}, the \textbf{goal} of this study is defined as: \emph{\textbf{analyze} the impact of \emph{Costmap 2D} configurations
\textbf{for the purpose of}
improving the energy efficiency
\textbf{with respect to}
ROS 2-based mobile robotic systems
\textbf{from the point of view} of
researchers and practitioners}. 
The above-mentioned goal is achieved by answering four research questions:

\noindent
\emph{\textbf{RQ1:}}
To what extent do configurations of \emph{Costmap 2D} impact the energy consumption of ROS 2 system? 

\noindent
\emph{\textbf{RQ2:}}
Which specific configuration parameters of the \emph{Costmap 2D} package contribute to energy efficiency during autonomous navigation?

\noindent
\emph{\textbf{RQ3:}}
How do workload conditions relate to \emph{Costmap 2D} configurations to influence energy consumption and system performance?

\noindent
\emph{\textbf{RQ4:}}
How can the findings from this controlled experiment inform good practices for researchers and practitioners?


%
%
%

\begin{table}[ht]
\centering
\vspace{-2mm}
\caption{Metrics Used in the Study}
\label{tab:metrics}
\begin{tabular}{p{0.9cm}p{1.6cm}p{5.0cm}}
\hline
\rowcolor[HTML]{AFAFAF}
\textbf{Name} & \textbf{Unit} & \textbf{Description} \\
\hline
\multicolumn{3}{c}{\textbf{Navigation Performance}} \\
\hline
\rowcolor{lightgray}
Time & seconds (s) & Duration for the robot to reach its goal. \\
Path Length & meters (m) & Total trajectory length from start to goal. \\
\rowcolor{lightgray}
Success & count (\#) & Trials in which the robot reaches the goal. \\
Recoveries & count (\#) & Instances where recovery behaviors are triggered to handle failures. \\
\hline
\multicolumn{3}{c}{\textbf{Energy Usage}} \\
\hline
\rowcolor{lightgray}
Energy & joules (J) & Overall energy consumed during navigation. \\
Power & mWatts (mW) & Average power demand during navigation. \\
\hline
\multicolumn{3}{c}{\textbf{Resource Usage}} \\
\hline
\rowcolor{lightgray}
CPU & hertz (Hz) & CPU cycles spent by Nav2 processes. \\
Memory & kBytes (KB) & Peak memory allocated by Nav2 processes. \\
\hline
\end{tabular}
\end{table}
%
%
%
%
Table~\ref{tab:metrics} summarizes the metrics used in the study.
We consider three categories of \textbf{experimental variables}:

\noindent
$\bullet$ \emph{Static variables}:
To ensure consistency and repeatability, certain aspects of the experimental setup are kept constant across all runs. These include the \emph{robot model} (TurtleBot 4, representative of real-world vacuum robots), the \emph{navigation stack} (same Nav2 installation), the \emph{hardware} and \emph{virtual machines} (VMs), the \emph{robot’s initial position} (mimicking a charging station), and the \emph{navigation goal}.

\noindent
$\bullet$ \emph{Independent variables}:
The factors under investigation comprise (i) the \emph{configurations} of the \texttt{Costmap 2D} package validated by two experts, (ii) the obstacle density (number of obstacles) within the environment, and (iii) the warehouse floor layout represented by two maps (small and large).

\noindent
$\bullet$ \emph{Dependent variables}:
These correspond to the metrics previously introduced, grouped into \emph{navigation performance}, \emph{resource usage}, and \emph{energy efficiency}.

\section{Experiment Execution}
\label{sec:execution}



All experiment runs are conducted on a workstation with these specifications: Linux Ubuntu \num{22.04} (kernel \texttt{6.8.0-65-generic}), \num{16}GB of RAM, a \texttt{12th Gen Intel® Core™ i7-12700F} processor with \num{20} cores (\num{8} performance and \num{12} efficiency cores), and an NVIDIA \texttt{GeForce RTX 3060} GPU. Under this configuration, no system resource was saturated during experimentation.
To guarantee isolation and reproducibility, we use four Docker (version \texttt{27.4.0}) containers: the \texttt{Gazebo} simulation environment, the \texttt{RViz} visualization tool, the \texttt{Nav2} navigation stack, and experiment orchestration services. For strict resource separation, both Docker and RViz containers were explicitly configured to run on the system’s GPU. The Gazebo real-time factor (RTF) was fixed at \num{1.0}, ensuring that the simulation advanced in synchrony with wall-clock time, thereby publishing sensor data at the same frequency as a physical robot.
Within Gazebo, we adopted the standard \emph{Turtlebot3 Waffle}\footnote{\url{https://github.com/ROBOTIS-GIT/turtlebot3}}
 model distributed with ROS 2 Humble, which is natively equipped with a 2D LiDAR scanner. In addition, we extended the robot model by integrating a 3D depth camera using the public \texttt{realsense\_gazebo\_plugin}\footnote{\url{https://github.com/pal-robotics/realsense_gazebo_plugin}}
 package.





\noindent
\textbf{\texttt{Costmap 2D} Configurations} --
Table~\ref{tab:costmap2d-params} presents the \num{18} Costmap 2D parameters considered in our experiment, which have been carefully selected based on ROS navigation tuning guides~\cite{Zheng2021, automaticaddison_ros_2024}.
These parameters and their possible values have been validated by two experts: (i) an academic researcher expert on Nav2 and (ii) the Nav2 project leader. 
An initial list of parameters and their suggested values were submitted to them via e-mail in June 2024 and refined based on their feedback.
Although some parameters are continuous in nature, we restricted them to a finite set of representative values, which follows typical ranges based on default values, covering \emph{low}, \emph{medium}, and \emph{high} values.

The full combination of all parameter values would result in more than 30 million configurations, making exhaustive experimentation impractical. 
We then use a pairwise combination strategy to generate a list of representative configurations for which experimentation is feasible. 
This technique, commonly employed in software testing~\cite{8030612, tai2002test}, is grounded in the observation that most software faults arise from interactions between pairs of parameters. 
Therefore, it should provide a representative coverage of the configuration space, possibly enabling the discovery of potential optimization opportunities.
As a result, we generated \num{20} representative configurations to be used in the experiment (see details in the replication package\footnote{\url{https://github.com/IntelAgir-Research-Group/ROS2-EE-Reconf}}).

The pairwise configuration generation considers predefined constraints, mostly implicit parameters related to plugins, and specified conditions highlighted by the experts. 
The most relevant constraint is the combination of plugins proposed by the experts for three navigation strategies:
i) \emph{avoiding dynamic obstacles by an alternative path calculated globally}, ii) \emph{avoiding dynamic obstacles by a global alternative path and local obstacle avoidance}, and iii) \emph{avoiding dynamic obstacles only locally}.
These combinations imply different trade-offs: from global replanning to hybrid global-local adaptation and local responsiveness.
For this, four plugins are used: \texttt{static\_layer} (\texttt{sl}), \texttt{obstacle\_layer} (\texttt{ol}), \texttt{voxel\_layer} (\texttt{vl}), and \texttt{inflation\_layer} (\texttt{il}).
The \texttt{sl} is responsible for representing immovable obstacles, which is always present in both local and global costmaps.
The \texttt{ol} incorporates real-time 2D sensor data to detected obstacles. 
The \texttt{vl} models obstacles in three dimensions, relying on an RGB-D camera sensor. 
Finally, the \texttt{il} enlarges the perceived size of obstacles by adding safety margins around them. 


\begin{table*}[htbp]
\centering
\caption{Costmap 2D parameter values used in the experiment}
\label{tab:costmap2d-params}
\begin{tabular}{lp{3.4cm}p{7cm}}
\hline
\rowcolor[HTML]{AFAFAF}
\textbf{Parameter Name} & \textbf{Possible Values} & \textbf{Description} \\
\hline
\texttt{update\_frequency} & \{1.0, 5.0, 10.0\} & Rate (Hz) at which the costmap is updated. \\
\rowcolor{lightgray}
\texttt{resolution} & \{0.1, 0.2\} & Resolution (in meters) of each cell in the costmap. \\
\rowcolor{white}
\texttt{width} & \{1.0, 2.0, 3.0\} & Width of the costmap in meters. \\
\rowcolor{lightgray}
\texttt{height} & \{1.0, 2.0, 3.0\} & Height of the costmap in meters. \\
\rowcolor{white}
\texttt{track\_unknown\_space} & \{true, false\} & If true, unknown space is tracked and visualized. \\
\rowcolor{lightgray}
\texttt{transform\_tolerance} & \{0.1, 0.5, 1.0\} & Tolerance time for TF transform lookup. \\
\rowcolor{white}
\texttt{plugin\_combination} & \{
\final{[ \texttt{sl},  (\texttt{sl},\texttt{ol},\texttt{il}) ],}
 & \multirow{3}{7cm}{Combination of layered costmap plugins used, divided in two sets (\texttt{controller} and \texttt{planner} plugins).} \\
& \final{[ (\texttt{sl},\texttt{ol}), (\texttt{sl},\texttt{il}) ],} & \\
& \final{[ (\texttt{sl}, \texttt{vl}), (\texttt{sl}, \texttt{ol}, \texttt{il}) ]} \} & \\

\rowcolor{lightgray}
\texttt{obstacle\_max\_range} & \{1.0, 2.0, 3.0\} & Max range to consider obstacles from scan data. \\
\rowcolor{white}
\texttt{raytrace\_max\_range} & \{1.0, 2.0, 3.0\} & Max range for raytracing clears from scan data. \\
\rowcolor{lightgray}
\texttt{voxel\_layer:z\_voxels} & \{5, 10, 15\} & Number of vertical voxels for 3D obstacle tracking. \\
\rowcolor{white}
\texttt{voxel\_layer:publish\_voxel\_map} & \{true, false\} & Whether to publish the 3D voxel map. \\
\rowcolor{lightgray}
\texttt{voxel\_layer:unknown\_threshold} & \{5, 10, 15\} & Threshold of unknown cells to consider a voxel unknown. \\
\rowcolor{white}
\texttt{voxel\_layer:mark\_threshold} & \{0, 1, 2\} & Number of marked voxels required to mark a grid cell. \\
\rowcolor{lightgray}
\texttt{inflation\_layer:inflation\_radius} & \{0.25, 0.5, 0.75\} & Distance around obstacles to inflate. \\
\rowcolor{white}
\texttt{inflation\_layer:cost\_scaling\_factor} & \{1.0, 2.0, 3.0\} & Rate at which inflation cost decreases with distance. \\
\rowcolor{lightgray}
\texttt{inflation\_layer:inflate\_unknown} & \{true, false\} & Inflate unknown regions as obstacles. \\
\rowcolor{white}
\texttt{inflation\_layer:inflate\_around\_unknown} & \{true, false\} & Inflate area around unknown space. \\
\rowcolor{lightgray}
\texttt{static\_layer:subscribe\_to\_updates} & \{true, false\} & If true, subscribes to external map updates. \\
\hline
\end{tabular}
\end{table*}

\noindent
\textbf{Experiment orchestration} --
We use Robot Runner (RR)~\cite{swanbornRobotRunnerTool2021} to orchestrate the experiment, handling the setup and execution of both Gazebo and Nav2, and performance/energy profiling. 
RR sequentially prepares the environment, launches the simulation with the robot and its configuration, starts profilers to monitor resource, energy, and navigation performance, drives the robotic mission, and finally cleans up the environment to ensure isolation for subsequent runs.
All the ROS 2 packages are set to use simulator time, which ensures that all nodes remain synchronized with the simulation clock instead of the system clock.
After each run, a \num{60}-second cooling-down period ensures that the machine returns to its previous resource usage levels.
During the experiment, we track resource usage, energy consumption, and navigation performance using customized \textbf{profilers}.
Energy consumption and CPU are measured via the RAPL interface through \emph{PowerJoular} profiler~\cite{noureddine2022powerjoular}. 
We measure the power of the whole machine, and of the two main ROS 2 nodes linked to \emph{Costmap 2D} configuration: \texttt{controller} and \texttt{planner}.
Memory is measured via \emph{psutil} Python library.
The performance metrics 
are measured by customized profilers that collect Nav2 navigation feedback\footnote {\url{https://docs.nav2.org/plugin_tutorials/docs/writing_new_navigator_plugin}}.


\section{Results}
\label{sec:results}

In this section, for a matter of space, we only illustrate the results of \emph{small} map.
However, all the data plotting is available on the replication package for further inspection.
Experiments with the \emph{small} map lasted a total of \(\approx33 \,
\text{h}\), with the overall duration given by 
\(T_{\text{total}}^{\textit{small}} = (d+cd) \times c \times o \times r\), where \(d\) is the average duration of a single run, \(cd\) the cool down period,
\(c\) the number of configurations, \(o\) the number of obstacle arrangements, and \(r\) the number of repetitions. 
Substituting the values, we obtain 
\(T_{\text{total}}^{\textit{small}} = (90+60) \times 20 \times 2 \times 20 \approx 120{,}000 \, \text{s} \equiv 33{,}33 \, \text{h}\).
Experiments with \emph{large} map lasted a total of \(\ 3 \, \text{days}\) (\(T_{\text{total}}^{\textit{large}} = (280+60) \times 20 \times 2 \times 20 \approx 272{,}000 \text{s} \equiv 3{.}14 \,\text{days}\)).

\begin{figure*}[htbp]
    \centering
    \begin{subfigure}[b]{0.32\textwidth}
        \centering
        \includegraphics[width=\linewidth]{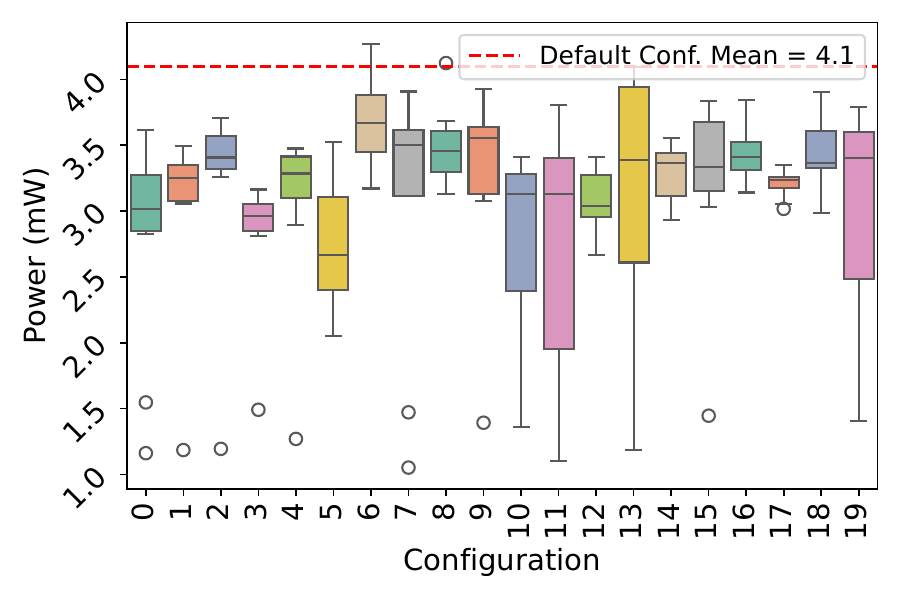}
        \vspace{-6mm}
        \caption{Power of \texttt{controller} node.}
        \label{fig:power_controller}
    \end{subfigure}
    \hfill
    \begin{subfigure}[b]{0.32\textwidth}
        \centering
        \includegraphics[width=\linewidth]{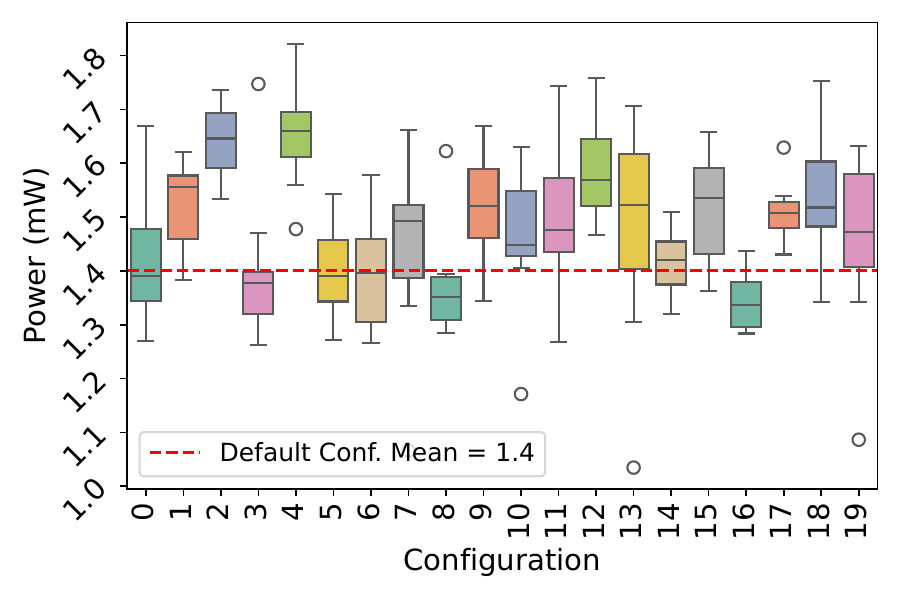}
        \vspace{-6mm}
        \caption{Power of \texttt{planner} node.}
        \label{fig:power_planner}
    \end{subfigure}
    \hfill
    \begin{subfigure}[b]{0.32\textwidth}
        \centering
        \includegraphics[width=\linewidth]{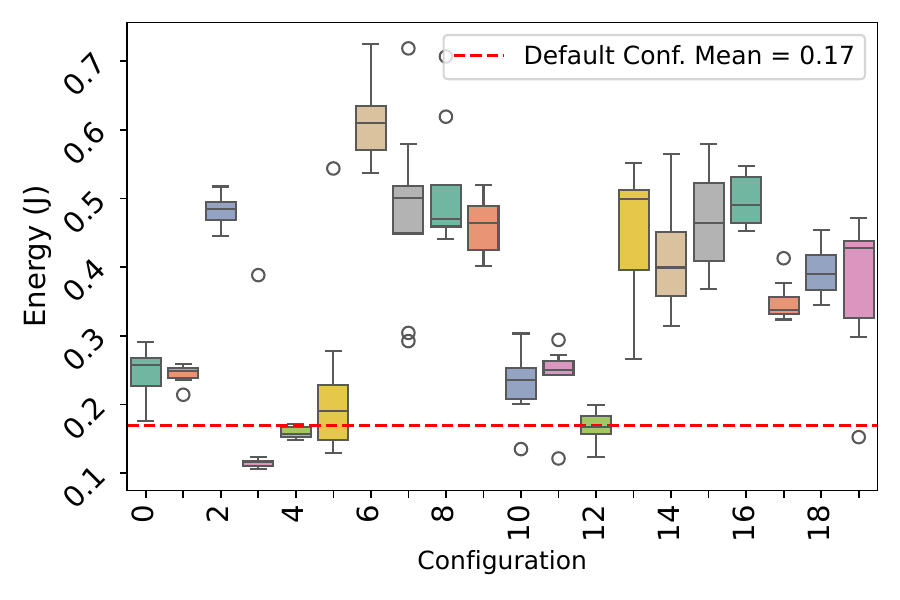}
        \vspace{-6mm}
        \caption{Energy consumption (both nodes).}
        \label{fig:controllerplannerenergy}
    \end{subfigure}
    \vspace{-1mm}
    \caption{Power and energy consumption per configuration for \texttt{controller} and \texttt{planner} nodes in \emph{small} map.}
    \label{fig:usage}
    \vspace{-0.5em}
\end{figure*}

\begin{figure}[htbp]
    \centering
    \begin{subfigure}[b]{0.23\textwidth}
        \centering
        \includegraphics[width=\linewidth]{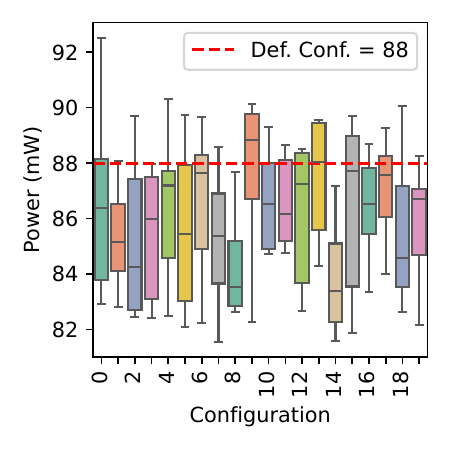}
        \vspace{-6mm}
        \caption{Machine Power.}
        \label{fig:machinepower}
    \end{subfigure}
    \hfill
    \begin{subfigure}[b]{0.23\textwidth}
        \centering
        \includegraphics[width=\linewidth]{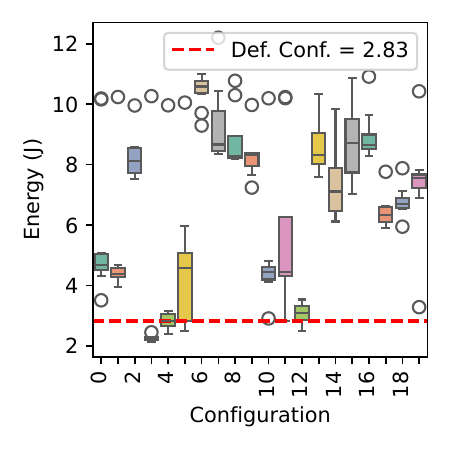}
        \vspace{-6mm}
        \caption{Machine Energy.}
        \label{fig:machineenergyuse}
    \end{subfigure}
    
    \caption{Power and energy consumption per configuration for the whole machine in the \emph{small} map.}
    \label{fig:machine-energy}
    \vspace{-1.5em}
\end{figure}

\begin{figure*}[hbtp]
    \centering
    \begin{subfigure}[b]{0.32\textwidth}
        \centering
        \includegraphics[width=\linewidth]{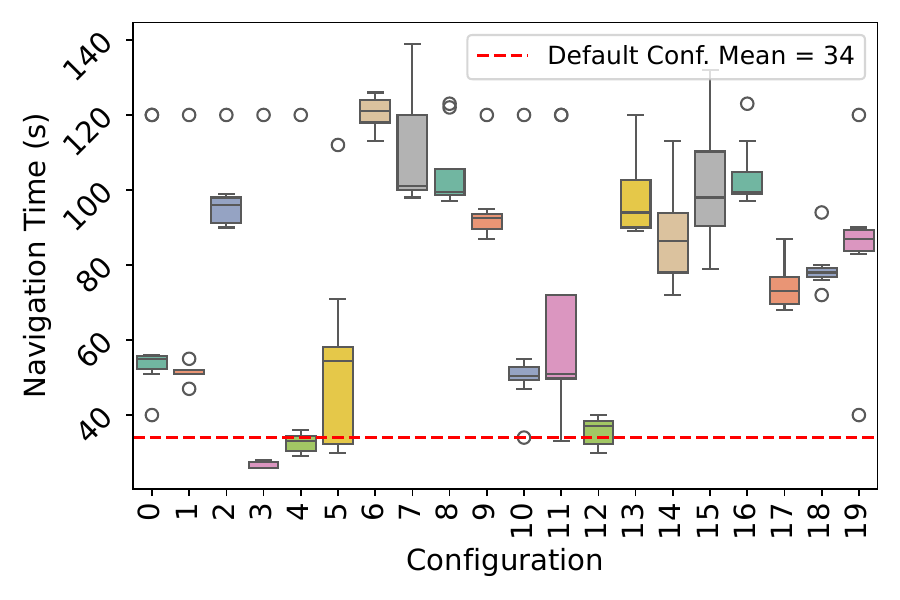}
        \vspace{-6mm}
        \caption{Navigation time.}
        \label{fig:navtime}
    \end{subfigure}
    \hfill
    \begin{subfigure}[b]{0.32\textwidth}
        \centering
        \includegraphics[width=\linewidth]{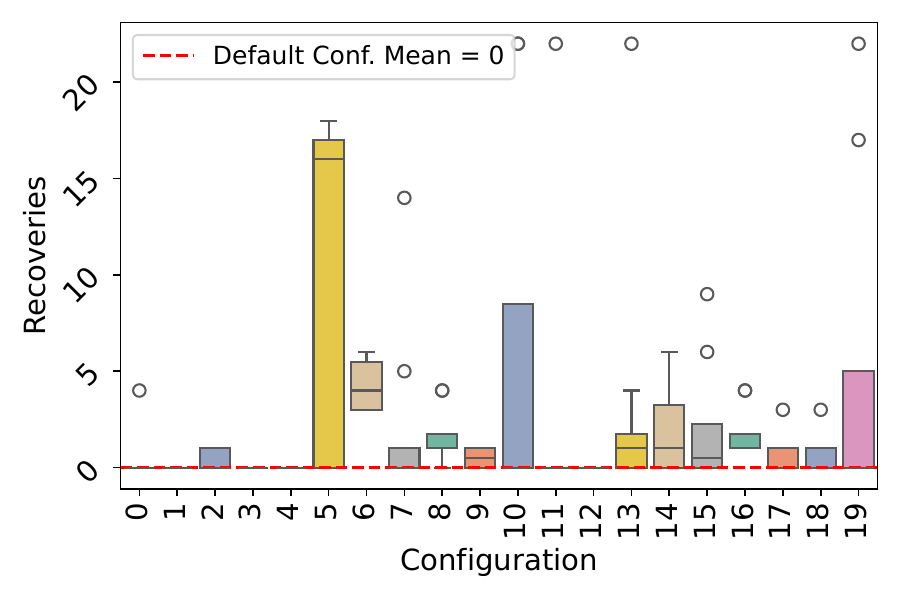}
        \vspace{-6mm}
        \caption{Navigation recoveries.}
        \label{fig:navrecoveries}
    \end{subfigure}
    \hfill
    \begin{subfigure}[b]{0.32\textwidth}
        \centering
        \includegraphics[width=\linewidth]{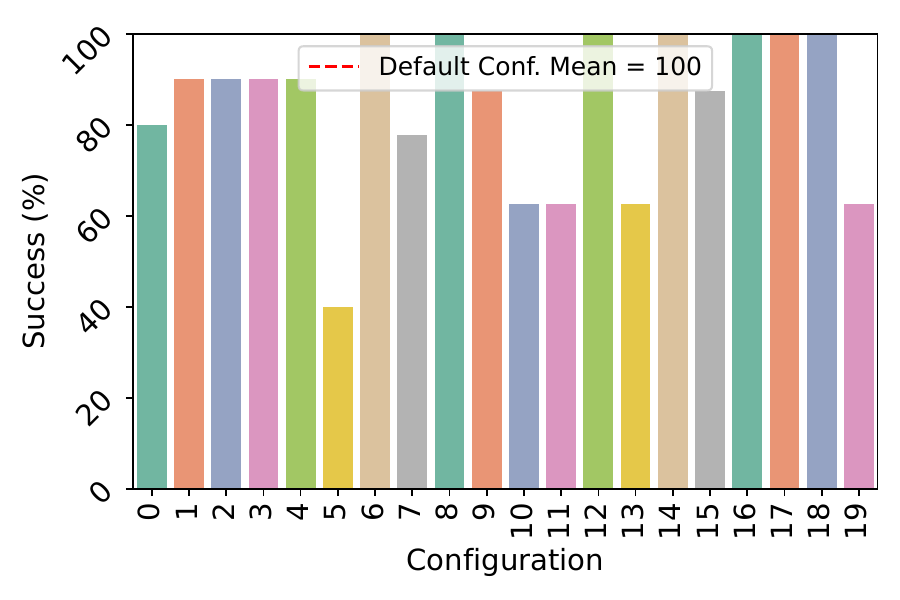}
        \vspace{-6mm}
        \caption{Navigation success.}
        \label{fig:navsuccess}
    \end{subfigure}
    \vspace{-1mm}
    \caption{Navigation performance per \texttt{Costmap\_2D} configuration in the \emph{small} map.}
    \vspace{-1em}
    \label{fig:navperf}
\end{figure*}


\subsection{Power and Energy Usage}

Figure~\ref{fig:usage} presents the distribution of \emph{power} and \emph{energy} usage of the \texttt{controller} and \texttt{planner} nodes in the \emph{small} map. 
For the \texttt{controller} node, all configurations consume less power than the \emph{default} setup. In contrast, for the \texttt{planner} node, only $c_2$ and $c_4$ exhibit higher power consumption than the \emph{default}. When considering aggregated \emph{energy} ($E = navigation\_{time} \times (controller\_{power} + planner\_{power})$, only three configurations $c_2$, $c_4$, and $c_{12}$ achieve lower overall energy consumption. This reduction is mainly due to extended navigation times in the proposed configurations, which lower frequencies and promote more cautious obstacle avoidance, ultimately forcing the robot to move at reduced speeds. A notable case is $c_4$: even with high planner power, its short navigation time results in low overall energy consumption.

Figure~\ref{fig:machine-energy} presents the distribution of CPU \emph{power} (Figure~\ref{fig:machinepower}) and \emph{energy} (Figure~\ref{fig:machineenergyuse}) consumption of the entire machine used to run the experiments in the \emph{small} map. 
Since the \texttt{Gazebo} and \texttt{Rviz} containers were executed on the GPU, they did not affect these CPU-based measurements. 
The results indicate that the configurations also directly affect the power and energy consumption of the entire Nav2 stack.
The \emph{power} of most configurations is lower than that of the \emph{default} configuration, with the exception of $c_9$ and $c_{13}$. However, when considering \emph{energy} consumption, only $c_3$ demonstrates higher efficiency, while $c_4$ and $c_{12}$ achieve values close to the \emph{default}. 

\begin{figure}[htbp]
    \centering
    \begin{subfigure}[b]{0.23\textwidth}
        \centering
        \includegraphics[width=\linewidth]{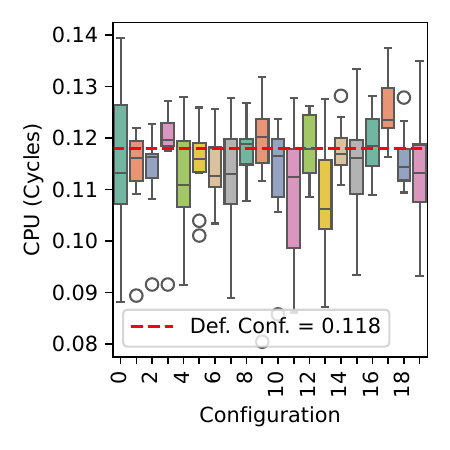}
        \vspace{-6mm}
        \caption{Machine CPU cycles.}
        \label{fig:machinecpu}
    \end{subfigure}
    \hfill
    \begin{subfigure}[b]{0.23\textwidth}
        \centering
        \includegraphics[width=\linewidth]{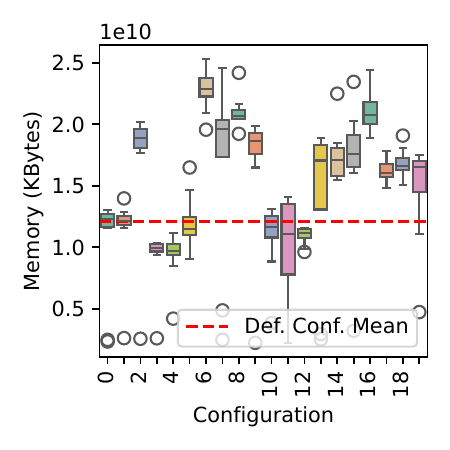}
        \vspace{-6mm}
        \caption{Machine memory usage.}
        \label{fig:machinemem}
    \end{subfigure}
    
    \caption{CPU and memory usage per configuration for the whole machine in the \emph{small} map.}
    \vspace{-1.5em}
    \label{fig:machinecpumen}
\end{figure}

\subsection{Resource Usage}

Figure~\ref{fig:machinecpumen} shows a variation in CPU usage across \emph{Costmap 2D} configurations. This behavior is expected, as it reflects the corresponding power consumption of the machine. However, Figures~\ref{fig:machinecpu} and \ref{fig:machinepower} are not consistently aligned. 
This discrepancy arises because CPU usage is typically reported in terms of active cycles, which capture logical activity that may result in different power usages, given that modern processors employ complex architectural mechanisms, such as instruction-level parallelism, cache techniques, and dynamic voltage and frequency scaling (DVFS)~\cite{6682055}.

Another observation is that memory usage also varies across \emph{Costmap 2D} configurations, with differences even more pronounced than those observed for CPU usage. In particular, the memory usage of $c_3$, $c_4$, and $c_{12}$ is consistently lower than that of the \emph{default} configuration, corroborating the reduced energy consumption observed for the \texttt{controller} and \texttt{planner} nodes (Figure~\ref{fig:controllerplannerenergy}). Since memory usage is not directly accounted for in our energy measurements, these results suggest that the impact of parameter configurations on the overall energy efficiency of the Nav2 stack may be even greater than indicated by CPU metrics alone.

\subsection{Navigation Performance}

Figure~\ref{fig:navtime} depicts \textbf{navigation time} across the \num{20} configurations in the \emph{small} map. The reduced energy consumption observed in $c_3$, $c_4$, and $c_{12}$ is primarily explained by their shorter navigation times. Several configurations ($c_0$, $c_1$, $c_6$, $c_{10}$, and $c_{11}$) result in intermediate navigation times of approximately \SI{50}{\second}, while the remaining configurations exceed \SI{70}{\second}. 

Considering the calculated \textbf{path length}, the default configuration performs worse than most of the proposed alternatives, with a mean distance of \num{6.49} meters. Only $c_1$ results in a longer mean path, while $c_2$ produces a path of identical length. In contrast, most configurations yield paths at least one meter shorter, largely due to factors such as restricted inflation layers around obstacles that encourage smoother or alternative trajectories. 
Path length does not seem to be determinant since $c_9$ more than doubles both navigation time and total energy consumption compared to the default, despite producing a path over one meter shorter, given the robot moves slower.

Figure~\ref{fig:navsuccess} presents the percentage of \textbf{navigation success} for each configuration. Configurations $c_6$, $c_8$, $c_{12}$, $c_{14}$, $c_{16}$, $c_{17}$, and $c_{18}$ achieve a $100\%$ success rate across all \num{20} repetitions, indicating safe and reliable navigation. 
In contrast, $c_5$, $c_{10}$, $c_{11}$, $c_{13}$, and $c_{19}$ proved problematic, with success rates below $80\%$. 
The causes of failures are analyzed in the discussion section. 
All remaining configurations completed more than $80\%$ of the trials successfully; 
however, as we explain in the discussion, the few unsuccessful runs in these cases are attributable to unexpected conditions in experiment orchestration, with no flaws in the configurations themselves.

Figure~\ref{fig:navrecoveries} illustrates the number of \textbf{recoveries} across each configuration. 
All configurations with low success rates ($< 80\%$) also exhibit a large number of recoveries, reinforcing the interpretation that they are problematic, particularly $c_5$. Two distinct behaviors emerge when comparing $c_6$ and $c_{11}$. Configuration $c_6$ achieves successful navigation in all runs with a high number of recoveries, suggesting that it relies heavily on corrective actions to complete its tasks. In contrast, $c_{11}$ reaches only a $60\%$ success rate despite requiring no recoveries, indicating that its failures result from conditions that could not be mitigated by recovery strategies.

\subsection{\emph{Large} Map Results}

A careful visual inspection of the \emph{large} map plots leads to several relevant observations. Unexpectedly, although the map is $\approx 10\times$ larger than the \emph{small} one, the mean CPU cycles, and consequently the measured power, are lower. 
This reduction is driven primarily by the \texttt{controller} node, whose power drops to roughly half, while the \texttt{planner} shows only a modest decrease. 
A plausible explanation is that the \emph{large} map affords wider corridors, keeping the robot farther from walls and obstacles and therefore triggering fewer avoidance maneuvers and micro-corrections at the controller level. 
Because path planning is typically invoked less frequently than obstacle avoidance, both \texttt{planner} nodes exhibit similar power demands, where in the \emph{large} map, the increased free space may also reduce path re-planning events.

Memory usage also decreases, although the reduction is more modest. 
This is expected since the overall map is larger, while the \texttt{controller} node still processes the same map area per CPU cycle across configurations. 
We identify four configurations with discrepant memory usage: $c_2$, $c_{17}$, $c_{18}$, and $c_{19}$. 
All of them employ the \texttt{voxel} plugin, 
which alone does not appear to drive high memory usage, 
with elevated consumption emerging when 
it is combined with additional factors. 
Specifically, these configurations also apply broader \emph{inflation} of obstacles and enable \emph{tracking of unknown space}, both of which expand the costmap representation.
Interestingly, under the \emph{small} map setting, $c_2$ stands out as one of the most memory-efficient configurations. 
Its main difference lies in the \texttt{cost\_scaling\_factor}, which is set to \num{3} in $c_2$, whereas in the other configurations it is fixed at \num{1}. 
This higher scaling factor makes the cost of distant obstacles decrease faster, so the inflated area is narrower, which reduces the number of cells stored and updated, contributing to lower memory usage.
It's particularly low memory usage on the \emph{small} map is also likely influenced by the limited extent of the map.

The navigation path length approximately triples in the \emph {large} map, which is also the case for the \emph{navigation time}.
This means the robot moves at a similar pace regardless of the map layout.
We also observe an unexpected behavior associated with excessive values of the \emph{inflation layer}. 
Such settings induced the robot to favor trajectories in close proximity to obstacles rather than exploiting the available free space. 
A representative case is $c_8$, where the navigation path was recalculated upon the introduction of the first obstacle. 
Instead of increasing clearance, the robot consistently detoured through narrow passages, for instance, between an obstacle and a pile of boxes. 
This behavior resulted in unnecessarily long and oscillatory trajectories, reflecting the tendency of the robot to remain near object boundaries. 







\subsection{Statistical Analysis}

We conducted a series of statistical tests for both the \emph{small} and \emph{large} maps. For each map, we examined the effect of (i) configuration and (ii) number of obstacles on the measured power of the \texttt{controller} and \texttt{planner} nodes, as well as on overall energy consumption. We also (iii) statistically compared the measurements for the two maps.
Shapiro--Wilk test~\cite{shapiro1968comparative}, in most cases, rejects the null hypothesis of normality, particularly for the \emph{small map} experiments, indicating non-normal distributions.
Given the frequent violations of normality, non-parametric tests were selected.

\subsubsection{Effects of configurations}
For the \emph{small map}, the test showed a strong and statistically significant effect of configuration on the \texttt{controller} power ($H{=}222.469$, $p{=}1.102{\times}10^{-36}$), the \texttt{planner} power ($H{=}237.752$, $p{=}9.255{\times}10^{-40}$), and the overall energy usage ($H{=}222.469$, $p{=}1.102{\times}10^{-36}$). 
For the \emph{large map}, configurations also had a significant influence, with smaller effect sizes: \texttt{controller} ($H{=}188.450$, $p{=}6.652{\times}10^{-30}$), \texttt{planner} ($H{=}138.150$, $p{=}4.118{\times}10^{-20}$), and overall energy ($H{=}188.450$, $p{=}6.652{\times}10^{-30}$).
Post-hoc tests revealed numerous significant pairwise differences among configurations in both maps, with specific parameterizations outperforming others.

\subsubsection{Effects of Obstacles}
Across both maps, obstacles did not yield statistically significant effects.
In the \emph{small map}, Kruskal--Wallis results were non-significant for the \texttt{controller} ($H{=}1.479$, $p{=}0.224$), \texttt{planner} ($H{=}0.026$, $p{=}0.871$), and overall energy ($H{=}1.479$, $p{=}0.224$). 
Similarly, in the \emph{large map}, results were non-significant for the \texttt{controller} ($H{=}1.720$, $p{=}0.190$), \texttt{planner} ($H{=}1.741$, $p{=}0.187$), and overall energy ($H{=}1.720$, $p{=}0.190$).
This indicates that, within the tested ranges, obstacle density had a limited impact compared to configuration choices.

\subsubsection{Cross-Run Comparisons (Small vs.\ Large Map)}
Direct comparisons between maps showed marked differences.
For the \texttt{controller}, Kruskal--Wallis indicated a very strong difference between maps ($H{=}683.666$, $p{=}1.066{\times}10^{-150}$), mirrored in overall energy ($H{=}683.666$, $p{=}1.066{\times}10^{-150}$). 
The \texttt{planner} also differed across maps, albeit to a lesser extent ($H{=}25.863$, $p{=}3.666{\times}10^{-7}$).
A two-way ANOVA (\emph{map} $\times$ \emph{obstacles}) corroborated these findings: a significant main effect of \emph{map} for \texttt{controller} ($F{=}690.28$, $p{=}2.76{\times}10^{-126}$), \texttt{planner} ($F{=}33.52$, $p{=}8.53{\times}10^{-9}$), and overall energy ($F{=}690.28$, $p{=}2.76{\times}10^{-126}$), with no significant main effect of \emph{obstacles} (all $p{>}0.17$) and no significant interaction (all $p{>}0.27$). 
Together with the post-hoc results, this confirms that map size/structure and configuration dominate energy/performance outcomes, while obstacle density does not modulate these effects in our setting.

\section{Discussion} 
\label{sec:discussion}


\subsection{Answering the Research Questions}


\noindent
\emph{\textbf{A--RQ1}}: 
Our experiments confirm that \emph{Costmap 2D} configurations have a substantial impact on the energy consumption of the Nav2 stack and, consequently, on the overall energy efficiency of ROS~2 systems. Parameter variations affect both the power profile of the planner and controller nodes as well as navigation time, which together determine total energy usage. Configurations such as $c_3$, $c_4$, and $c_{12}$ consistently achieved lower energy consumption than the default setup, primarily by reducing navigation time and lowering node-level power. These configurations also exhibited reduced memory usage, an aspect not directly captured in our power and energy measurements, which may represent an additional dimension of energy efficiency improvement. 
Overall, these findings demonstrate that \emph{the energy efficiency of the Nav2 stack can be improved through deliberate configuration tuning}.


\noindent
\emph{\textbf{A--RQ2}}:
The analysis reveals that parameter tuning has a greater influence than plugin combinations. 
Three parameters stand out: \emph{resolution} -- 
a moderate value ($0.1 \, m$) reduced computational load without compromising navigation reliability, while a rough setting ($0.2 \, m$) consistently caused unsuccessful navigation;
\emph{update frequency} -- higher frequencies ($\geq5 \, Hz$) improved responsiveness and lowered \texttt{controller} power, whereas very low values ($1 \, Hz$) increased \texttt{planner} power due to more intensive replanning;
\emph{inflation} and \emph{unknown-space handling} -- disabling \texttt{inflate\_around\_unknown} and \texttt{track\_unknown\_space} parameters reduced unnecessary complexity in costmaps, shortening navigation times.



\noindent
\emph{\textbf{A--RQ3}}:
Map size and layout are directly related to how \emph{Costmap 2D} configurations affect energy consumption and Nav2 performance. 
In the \emph{large} map, although the environment is $\approx 10\times$ larger, the mean CPU cycles and power are lower than in the \emph{small} map. 
This is mainly linked to the \texttt{controller} node, which in wider corridors reduces the micro-corrections in the robot's path and faces farther objects.
Contrastingly, narrow environments force frequent avoidance and re-planning, making configuration choices more critical: combinations of \texttt{voxel} layers, broad obstacle inflation, and \texttt{track\_unknown\_space} increase memory usage, while higher \texttt{cost\_scaling\_factor} narrows inflated regions, leading to improved efficiency. 
Thus, \emph{performance-energy trade-offs must also consider different environmental conditions when setting the Nav2 parameterization}.
The obstacle arrangements used in our experiments did not produce statistically significant differences in the measurements, even though they were deterministically placed to trigger path replanning. 
Nevertheless, experiments involving a larger number of obstacles, arranged in more dynamic ways, could affect the measurements differently, requiring further experimentation.


\noindent
\emph{\textbf{A--RQ4}}:
The findings suggest practical guidelines: i) parameter tuning should be prioritized over plugin selection for energy efficiency;
ii) configurations must be validated under realistic workloads, as settings that appear efficient in simple environments may fail in more complex and large maps; iii) navigation time should be closely monitored, as it is the most influential variable for energy consumption; and
iv) results underscore the need to integrate energy measurement into the ROS development workflow, enabling practitioners to detect inefficient parameter settings early and researchers to build systematic knowledge for energy-aware robotics software.

\subsection{Practical Observations}

\noindent
\textbf{Parameter tuning matters more than the combination of plugins} --
Analyzing the configurations that yield lower power and energy consumption, we find no consistent pattern in the choice of local and global plugins, suggesting they are not decisive factors. 
This observation is particularly worthy of note, as certain plugins, such as the \texttt{voxel\_layer}, are expected to be computationally demanding due to three-dimensional representations of the environment.

The configurations that outperform the \emph{default} in terms of energy efficiency ($c_3$, $c_4$, and $c_{12}$) share three characteristics: shorter \emph{navigation times}, lower power usage in the \texttt{planner} and/or \texttt{controller} nodes, and the absence of recovery behaviors.
These configurations also follow a common parameter pattern: \texttt{resolution} set to \num{0.1}, \texttt{inflate\_around\_unknown} disabled, and a high \texttt{update\_frequency} ($\geq 5$). 
In contrast, the \emph{default} configuration uses a finer \texttt{resolution} of \num{0.05}, which emerges as the main differentiating factor. 
Lower \texttt{resolution} reduces the number of grid cells, simplifying the costmap and lowering the computational load, which in turn decreases power consumption. 
However, setting the \texttt{resolution} too high (\num{0.2}) leads to a large number of unsuccessful navigation attempts, highlighting the need for careful calibration of this parameter.



\noindent
\textbf{Navigation time is variable and determinant} --
In the graphs of Figure~\ref{fig:usage}, navigation time emerges as one of the most variable dependent measures. 
This variable is critical for estimating the total energy consumption of a robotic mission, as longer operating times inherently lead to greater energy usage.
Among the configurations with the longest navigation times ($c_2$, $c_6$, $c_7$, $c_8$, and $c_{16}$), a consistent pattern can be observed: \texttt{inflation\_radius} is set to \texttt{0.5}, and \texttt{track\_unknown\_space} is enabled. 
A large inflation radius (\texttt{0.5}) reduces available free space by enlarging obstacles, forcing detours, while 
enabling \texttt{track\_unknown\_space} further restricts motion by treating unexplored areas as unsafe. 
In the \emph{large} map, it was observed that a large inflation radius forces the robot to navigate close to obstacles, avoiding open spaces, which increases the navigated distance and time.
For the configuration with the shortest navigation time ($c_3$), \texttt{inflation\_radius} is set with value \num{0.35} and \texttt{track\_unknown\_space} is disabled, confirming them as determining factors in reducing navigation time. However, since those parameters are equally set in configurations $c_9$, which exhibit longer navigation time.
The main difference between $c_3$ and $c_9$ is the \texttt{transform\_tolerance}=\num{0.1} (vs. \num{0.5}), making the costmap wait longer for buffered transforms in $c_9$, accumulating pauses during navigation.
So, navigation time seems to result from a combination of those factors.

\noindent
\textbf{Reasons behind unsuccessful navigation} -- An analysis of the configurations leading to unsuccessful navigation reveals that some configurations performed particularly poorly ($<80\%$), with failures primarily due to crashes where the robot collided with obstacles, especially at sharp corners. 
These crashes are likely explained by a combination of \texttt{Costmap 2D} \emph{resolutions}, \emph{restrictive sensing ranges}, and \emph{limited local planning space}. With a resolution of \num{0.2}~m, the local costmap lacks sufficient granularity to represent narrow passages or fine environmental details, causing the robot to misinterpret free space and obstacle boundaries~\cite{Zheng2021}. 
Furthermore, restricting obstacle detection to only \num{1}~m results in late obstacle recognition, reducing the system’s reaction time. 
Finally, the absence of a local inflation layer eliminates smooth navigation gradients, increasing the likelihood of the robot hitting obstacles. 

For configurations with a success rate above \num{80}\% that still resulted in at least one unsuccessful navigation, we did not observe any crashes.  
Upon closer inspection of the experimental data, we found that these runs terminated after only \num{20} seconds, with no navigation performance collected. 
This deterministic behavior occurs when the robot fails to receive the navigation goal within \num{15} seconds, a situation that may arise from transient or unstable conditions in the experimental environment rather than from the navigation process itself. 
Since these cases are artifacts of the experimental setup and do not reflect the robot’s actual navigation capability, they can be safely treated as outliers.  
These outliers represent less than \num{10}\% of the collected runs, and their removal does not distort the underlying data assumptions. 

\noindent
\textbf{Configurations are not straightforward} --
%
%
%
%
Configuring the \emph{Costmap 2D} parameters in Nav2 is not straightforward when the goal is to balance performance and energy consumption. 
Our results show that the \emph{default} configuration is not necessarily ideal, even for simple missions, as it reflects general-purpose trade-offs rather than mission-specific optimizations. 
Moreover, performance and energy efficiency emerge from the interplay of multiple parameters rather than from any single setting, which becomes even more critical in larger or dynamic environments where replanning and obstacle avoidance are more frequent. 
Therefore, there is a need for adaptive configuration strategies: while static tuning is useful for controlled experiments, real deployments require mechanisms that reconfigure parameters at runtime. 



\noindent
\final{
\textbf{Threats to external validity} -- Our experimental evaluation is limited to the Nav2 navigation stack, specifically focusing on the \emph{Costmap~2D} package.
Moreover, all experiments were conducted using a single robot model, with fixed sensors and hardware settings. 
Future work will replicate the experiments across diverse robot platforms and extend the analysis to additional ROS~2 packages. 
}
\section{Conclusion}
\label{sec:conclusion}

This paper empirically demonstrates that ROS 2 \emph{Costmap 2D} configurations significantly influence the energy efficiency of mobile robot navigation. Through controlled experiments across different maps and obstacle layouts, we showed that parameter tuning, particularly \emph{resolution}, \emph{update frequency}, and \emph{inflation settings}, has a stronger impact on energy consumption than plugin combinations. 
Configurations such as $c_3$, $c_4$, and $c_{12}$ consistently reduced \emph{navigation time}, lowered \emph{power} demands of \texttt{controller} and \texttt{planner} nodes, and decreased \emph{memory} usage, thus outperforming the \emph{default} setup. 
Our findings highlight that \textbf{configuration tuning is essential to achieve efficient navigation in ROS 2 systems}. 



\bibliographystyle{IEEEtran}
\bibliography{IEEEabrv,bibliography}


%

\end{document}